\documentclass[final,5p,times,twocolumn,authoryear]{elsarticle}

\usepackage[utf8]{inputenc}
\usepackage[T1]{fontenc}
\usepackage{graphicx}
\usepackage{amsmath, amssymb, amsfonts}
\usepackage{tabularx}
\usepackage{booktabs}
\usepackage{caption}
\journal{}

\begin{document}
\begin{frontmatter}

\title{On the Disagreement in Perturbation-based xAI - Benchmarking Perturbation Choices for Flood Detection from SAR Images} 
\author[dlr,tum]{Anastasia~Schlegel}
\author[dlr]{Ronny~Hänsch}

\affiliation[dlr]{
    organization={Microwaves and Radar Institute, German Aerospace Center (DLR)},
    city={Weßling},
    postcode={82234},
    country={Germany}
}

\affiliation[tum]{
    organization={Chair of Remote Sensing Technology, Department of Aerospace and Geodesy, Technical University of Munich},
    city={Munich},
    postcode={80333},
    country={Germany}
}

\begin{abstract}
Perturbation-based xAI methods are widely used to analyze the behavior and predictions of deep learning models. 
By altering input regions and measuring the resulting changes in class probabilities with respect to the original image, they assign relevance scores and generate heatmaps that reflect each region's contribution to the prediction.
Despite their apparent simplicity, however, perturbation-based methods are sensitive to parameter choices.
In this work, we focus on two key parameters of the perturbation pipeline, namely the patch geometry, including the size and shape of the perturbed regions, and the perturbation type, defined by the replacement scheme.
Grounded in the use case of flood detection from Synthetic Aperture Radar imagery, we conduct a comprehensive investigation of how relevance estimation changes under different perturbation settings. 
Beyond visual inspection of the resulting relevance maps, we evaluate their consistency across perturbation strategies and their faithfulness to the model's reasoning.
We demonstrate how different perturbation choices can steer the resulting relevance maps, yielding ambiguous and even contradictory explanations. 
Our findings emphasize the importance of methodological settings in perturbation-based xAI.
They underscore the need to carefully inspect and evaluate perturbation choices and to treat them as an integral part when interpreting explanations, ensuring a robust understanding of both the explanations and model predictions.
\end{abstract}


\end{frontmatter}

\section{Motivation and Introduction}
In Earth observation (EO) applications deep learning and in particular convolutional networks (ConvNets) have become increasingly prevalent in recent years \citep{xai_roscher_20, xai_campsvalls_21}.
The capacity to learn complex representations drives their strong performance, yet also makes it difficult to understand how they arrive at their decisions.
It is often unclear whether a correct prediction is driven by salient features or by spurious artifacts, giving ConvNets the reputation of being black boxes~\citep{xai_rudin_19}.
This lack of transparency limits the interpretability of predictions and, in turn, constrains trust, reliability, and the verifiability of model decisions, which are particularly required in critical settings.
Spurious correlations, dataset artifacts, label noise, class imbalance, and other factors can subtly bias the learned decision rules. This motivates efforts to inspect the inherent drivers of a model’s prediction \citep{xai_samek_19}.

In response to this need for transparency, research on explainable AI (xAI) has emerged aiming at making model behavior and model predictions interpretable to humans \citep{xai_doshivelez_18}.
The rapid growth of xAI research in the recent years has produced a broad landscape of explanation techniques, with attribution methods forming a central group of post-hoc approaches \citep{saliency, guided_backprop, occlusion, lrp, cam, lime, deeplift, int_grad, gradcam, shap}.
Attribution methods provide local interpretability when applied to an already trained model by explaining individual predictions for a given input instance \citep{ancona_19_book}. 
They assign a scalar attribution value, often referred to as \textit{relevance} or \textit{contribution}, to each input feature with respect to a target output. 
In the case of image data these scores are typically visualized as heatmaps (saliency maps, attribution maps, relevance maps) that indicate where evidence and counter evidence for the target accumulate. 
Within attribution methods, perturbation-based approaches \citep{occlusion, lime, pred_differnce, meaningful_pert, rise, mfpp} estimate relevance by systematically altering parts of the input and measuring how the model output changes. 
Regions are masked, removed, or reshaped, and the effect on the prediction is interpreted as the influence of the perturbed features. 
Existing perturbation methods differ by their underlying perturbation technique.
Early work on occlusion is introduced by \citet{occlusion} by moving a gray square mask across the image, replacing one patch at a time and measuring how the class score changes. 
Similarly, Prediction Difference Analysis (PDA) \citep{pred_differnce} determines relevance by removing patches in a sliding window manner and replacing the regions by either sampling patches from other images at the same spatial location or sampling values from a Normal distribution based on the surrounding neighborhood.
Instead of replacing image regions with a fixed value, RISE \citep{rise} perturbs the input by multiplying it element-wise with randomly sampled soft masks, containing values in $[0,1]$, which attenuates pixel intensities. 
To form a saliency map, RISE aggregates the random masks weighted by the model's target class score on the corresponding masked inputs.
MFPP~\citep{mfpp} follows the same sampling and aggregation principle of RISE, but replaces grid-like masks with multiscale segmentation fragments. 
LIME \citep{lime} perturbs the input by randomly removing subsets of superpixels, typically by replacing the masked regions with a fixed gray baseline value, and uses these samples with local weighting to fit a sparse linear surrogate that approximates the network.
Rather than relying on randomly sampled masks, Meaningful Perturbations \citep{meaningful_pert} learn a free form mask.
Using a second network, masking is optimized, such that the image changes as little as possible while the target class score strongly decreases. The learned mask can control different deletion operators such as constant value replacement, noise replacement, or blurring.

Observing the input-output relationship is an intuitive way to probe model behavior, making perturbation-based methods widely used among existing xAI techniques \citep{eval_ivanovs_21}.
Moreover, they are typically model-agnostic, requiring no access to the model’s inner workings.
At the same time, perturbation-based explanations can become computationally exhaustive as the number of forward passes through the model increases with the number input features.
More critically, prior work provides first indications that perturbation-based explanations tend to be sensitive to the perturbation design \citep{ancona_19_book}.
In studies examining individual design choices, variations in region geometry, feature grouping, or the replacement scheme have been shown to alter the estimated relevance patterns yielding explanations that differ across perturbation settings.
\citet{krishna_24} describe this as the \textit{disagreement of explanations}.
However, these observations are typically drawn from specific perturbation variants rather than spanning the broader design space, motivating a systematic characterization of how perturbation choices shape the resulting attributions.

\begin{figure}[ht]
  \centering
  \includegraphics[width=\linewidth]{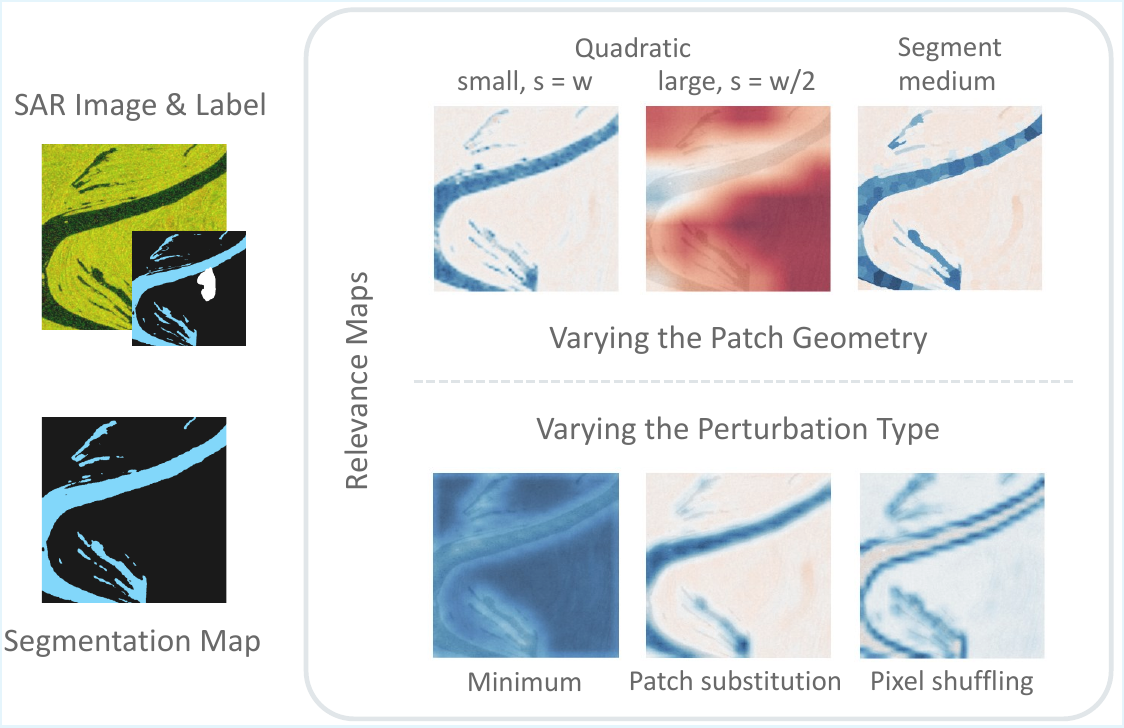}
  \caption{Disagreement of perturbation-based explanations for a sample SAR image. Left: SAR image with its reference label and the predicted segmentation map. Right: Relevance maps obtained under chosen perturbation settings, where blue indicates important regions and red disturbing ones. The top row shows variations of the perturbation patch geometry using the mean of the background class as the replacement value: small non-overlapping, large overlapping square windows (where  w = window size and s = stride), and medium-sized superpixels. The bottom row shows variations of the perturbation type using medium-sized overlapping square windows: channel minimum, patch substitution, and pixel shuffling.}  
  \label{fig:intro}
\end{figure}

To address this gap, we systematically investigate how perturbation design choices influence relevance estimation in perturbation-based attribution.
Specifically, we conduct a comprehensive analysis of two key parameters: i) the perturbation patch geometry, including size and shape, and ii) the perturbation type, given by the replacement scheme, and show how these choices affect the resulting relevance maps.
Figure~\ref{fig:intro} illustrates how the relevance maps for a sample image differ under varying perturbation choices.
Beyond qualitative inspection, we compare spatial relevance distribution patterns across perturbation variants to assess whether the model consistently relies on similar input features or whether its focus shifts, helping to distinguish robust patterns from masking artifacts.
Finally, we evaluate the faithfulness of explanations obtained under different perturbation designs, providing a structured view of explanation behavior across perturbation strategies and scene types.

Our analysis is grounded in the critical use case of flood detection from Synthetic Aperture Radar (SAR) imagery.
Globally operating spaceborne SAR enables rapid, repeated observations of flooded regions, and owing to its active sensing principle, it provides acquisitions that are largely unaffected by cloud cover and heavy rainfall that commonly accompany flood events \citep{floods_alfieri}.
While SAR image analysis increasingly adopts models and tools originally developed for natural images \citep{ai4eo_tuia_25}, transferring perturbation-based xAI approaches to SAR data requires particular care, as replacement values can carry physical meaning.
Standard gray window perturbations, for instance, may introduce artifacts and distort the inferred relevance patterns.
We therefore benchmark perturbation choices with particular attention to SAR-specific constraints.
\section{Disagreement in Attribution Methods}
By offering a comprehensive analysis of perturbation choices and their effects on feature attributions, we contribute to a line of research on the disagreement problem in attribution methods.
Disagreement refers to the variability in attribution results for the same prediction, arising from the choice of attribution method as well as from the choice of parameters within the same method.
This variability constraints the interpretation and reliability of explanations, motivating a rigorous examination of the methodological decisions.
In this section, we contextualize our contribution by outlining related works that characterize disagreement across attribution methods and parameter choices.

Across methods, \citet{krishna_24} formalize disagreement by distinguishing whether explainers identify the same critical features, how they order these features by importance, and whether features contribute positively or negatively to the prediction.
To make explanations comparable, they introduce several metrics that quantify disagreement between two explainers for the same instance.
While their metrics do not indicate a preference of one explainer over another, they characterize the extent of agreement and provide guidance for interpreting it.
Strong disagreement on a given instance is a warning signal that a single explanation should not be trusted without additional checks, and high agreement can still be jointly wrong. 
They therefore argue that the choice of the explainer should be justified through evaluation rather than be selected ad-hoc. 

Disagreement can further emerge within a single attribution method when parameter settings vary.
Especially in baseline-based methods, feature contributions are quantified by comparing the input to a reference (baseline), whose choice can strongly shape the resulting explanations.
Discussing the importance of baseline choice in Integrated Gradients, \citet{sturmfels_20} frame baseline selection through the concept of missingness.
In this view, the baseline should represent the absence of features rather than an arbitrary default.
They note that the black image baseline, although common in natural image settings, can be a poor proxy for absence in certain domains and applications.
Because attributions depend on the difference to the baseline, such constant value baselines generally are prone to unintended biases, as regions whose values match the baseline can receive low attribution scores even when they are important. 
By discussing alternative baseline constructions and how they shape the resulting attributions, their work provides a conceptual foundation for treating baseline choice as an integral part of interpreting explanations.
Beyond motivating baseline choice conceptually, \citet{mamalakis_23} propose aligning baseline selection with the kind of insights one aims to extract.
In the context of climate modeling, they compare Integrated Gradients and SHAP under multiple baseline definitions.
Varying the baseline can reveal complementary aspects of the learned mapping and thereby support different scientific questions.
They argue that baseline choice is often overlooked in geoscientific literature and therefore suggest selecting baselines intentionally and in line with the explanatory objective, to be aware of how baselines can steer attribution maps.

In perturbation-based approaches, which form a central class of baseline-based attribution methods, disagreement has been noted in \citet{ancona_18} and \citet{ancona_19_book}. 
For the original occlusion framework, they demonstrate how masking with a gray square can shift the focus across heatmaps as the occlusion window size changes, suggesting that different feature aggregations induce different explanations.
Occlusion trades granularity for computational feasibility, making the resulting explanations sensitive to the perturbation design, with the attribution shifting as the occlusion settings change.
While their work provides initial observations on how the perturbation operator shapes explanations, it falls short of a systematic analysis and evaluation of these design choices. 
This gap motivates a deeper investigation of the perturbation design to assess which choices yield stable and faithful attributions.

A small number of studies address this question by analyzing perturbation design choices more rigorously. 
These works most closely relate to our contribution.
\citet{pihlgren_25} present a comprehensive investigation of perturbation parameters. 
Building on the perturbation pipeline in RISE, they evaluate how choices in segmentation, sampling, perturbation, and attribution affect the resulting explanations. 
Specifically, they vary between grid and superpixel segments, apply a segment-wise constant value perturbation that fades toward segment edges, sweep the sampling scheme and sample size, and compare multiple attribution methods.
They assess these combinations using the Symmetric Relevance Gain metric, which balances perturbing most influential regions first and least influential regions last to reduce sensitivity to a single deletion ordering. 
Their results suggest that segmentation and sampling choices can have a stronger impact on the explanations, than attribution. 
This aligns with our motivation to understand which parts of the perturbation pipeline drive explanation variability. 
While their study provides a pipeline level view, systematically assessing how each design choice in the perturbation framework shapes the explanation, our contribution is more specific in scope and deeper in analysis, focusing on selected parameters.
A complementary perspective on disagreement in perturbation-based attribution is offered by \citet{bluecher_24}, who focus on distribution shifts induced by occlusion.
They argue that occluded samples are artificial, in the sense that occlusions can create inputs that fall outside the training distribution, pushing the model into regions where it has not learned reliable associations.
Consequently, prediction changes may reflect perturbation artifacts rather than the removal of genuine class evidence, making explanations unreliable.
Through this lens, they compare occlusion strategies and introduce the R-OMS score as a measure of reliability.
It is defined by the model’s confidence in the reference class when evaluated on the occluded input. 
Intuitively, out-of-distribution occlusions tend to suppress the model’s confidence in the reference class, indicating artifact driven effects, whereas more plausible occlusions preserve it. 
Using the R-OMS score, they present a comprehensive analysis of the reliability of perturbation attributions, considering varying inpainting styles and segmentation types across different models. 
Our work is driven by the same motivation in that it offers a comprehensive analysis of perturbation strategies, yet the two studies emphasize different aspects.
Although we acknowledge that occlusions can introduce artifacts, this does not form the viewpoint our analysis. 
Instead, we inspect the effects induced by different perturbation designs and discuss how these effects may relate to perturbation artifacts.
Accordingly, we do not aim to rank occlusion strategies by reliability.
Rather, we assess whether the induced effects and resulting attributions are consistent across perturbation strategies, and evaluate the faithfulness of explanations with respect to the model.

\section{Explainability in Flood Mapping}

\begin{figure*}[t]   
  \centering
  \includegraphics[width=\textwidth]{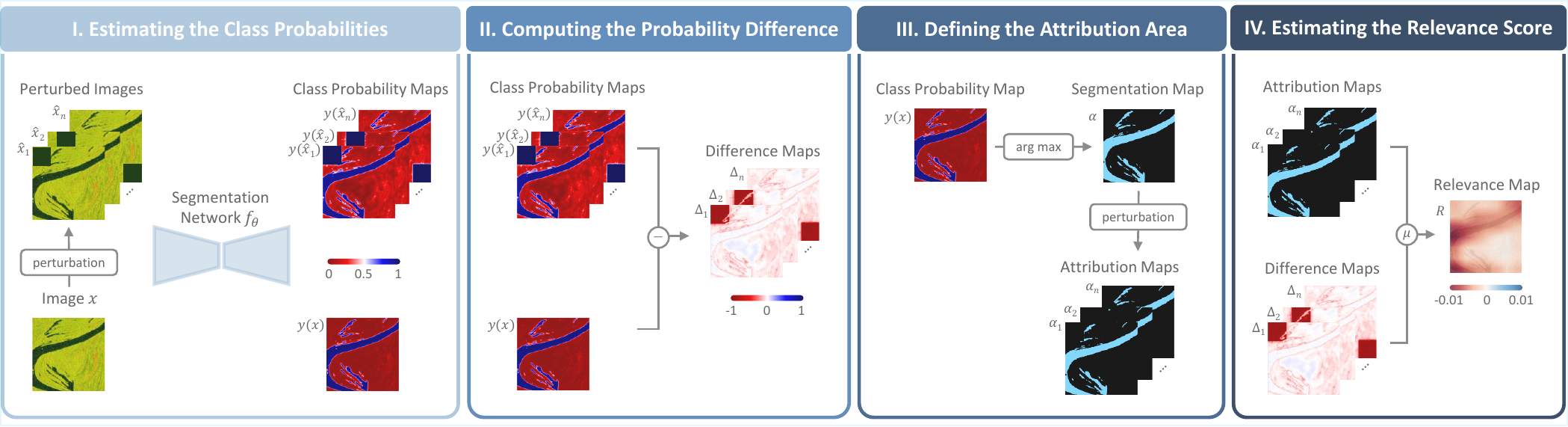}
  \caption{Perturbation-based relevance estimation approach: I. Create locally perturbed variants of an input image and infer pixel-wise class probability maps. II. Derive perturbation-induced probability difference maps. III. Determine the attribution area from the discrete segmentation of the input image. IV. Compute patch-wise relevance scores by averaging differences over the attribution area and project the scores back to the image grid to form a dense relevance map.}
  \label{fig:method}
\end{figure*}

Our work extends the research landscape on explainability in the context of flood detection.
Existing studies largely use xAI as a diagnostic tool, by leveraging post-hoc explanations to gain insight into model behavior and to support qualitative validation of flood predictions. 
\citet{sota_sanderson} present an explainable framework for flood inundation mapping that combines multimodal segmentation with post-hoc explanations. 
Their network, XFIMNet, is inspired by the U-Net and is extended to handle both SAR and multispectral inputs. 
It is followed by a Grad-CAM module, that identifies input regions driving the inundation prediction.
In their work, explainability is leveraged to compare model behavior across architectural variants and input configurations, with Grad-CAM applied in line with best practice by targeting the deepest convolutional layer.
Similarly, \citet{sota_chen} employ attribution analysis to validate model reasoning in a surface-water detection framework and to compare the effectiveness of different feature extractors.
By combining local and global attributions, they quantify and visualize feature contributions, and introduce indices that enable quantitative comparison of heatmaps across backbone networks.
A related direction is taken by \citet{sota_bathe}, who extend a flood detection pipeline with SHAP to support interpretation and validation of model decisions.
After training their dilated-convolution-based ConvExNet, they estimate Shapley values per input feature to capture both positive and negative contributions to the flood decision.
Using NDWI images as input, the resulting attribution maps indicate which NDWI pixels support or suppress the predicted flood class. 
Against these studies, our work shifts the focus from producing explanations to assessing the explanation quality, thereby promoting a more careful and justified use of xAI methods. 

\section{Perturbation-based Relevance Estimation}\label{sec:method}

Perturbation-based explanation methods provide insights into a model's decision-making by systematically altering input data and observing how the modifications affect the model's output.
In our use case, flood detection is formulated as a semantic segmentation problem discriminating water from background.
For an input image $x$, the model $f_{\theta}$ produces per-pixel logits, which are converted into pixel-level class posterior probabilities $y(x)$ using softmax.
To quantify the influence of an input region on a user-defined attribution area (e.g. the predicted area of a class), we compare these class posterior probabilities within the attribution area before and after the perturbation.
If perturbing an input region decreases the predicted class posterior, this indicates that this input region is crucial for the model's decision-making and that the model relies on that region when forming its prediction.
Conversely, if altering an input part increases the predicted class posterior, this suggests a misleading or distracting influence on the model's decision-making. 
Input perturbation supports identifying input regions the model uses as evidence and counter-evidence when forming its prediction. 
It allows verifying if the model's decision is mainly based on the object itself or if it is influenced by the surrounding context. 
Figure~\ref{fig:method} presents an overview of the perturbation-based relevance estimation approach in four steps which are discussed in more detail below.

\renewcommand\tabularxcolumn[1]{m{#1}}
\newcolumntype{L}[1]{>{\raggedright\arraybackslash}m{#1}}
\begin{table*}[t]
\centering
\small
\caption{Perturbation types considered in this study.}
\label{tab:ptype}
\renewcommand{\arraystretch}{1.15} 
\setlength{\tabcolsep}{1pt}
\begin{tabularx}{\linewidth}{@{} L{0.31\linewidth} L{0.14\linewidth} >{\raggedright\arraybackslash}X @{}}
\toprule
\textbf{Perturbation Type} & \textbf{Example} & \textbf{Description}\\
\midrule
\addlinespace[1.8mm]
\textbf{Channel-wise minimum}             & \includegraphics[width=1.5cm]{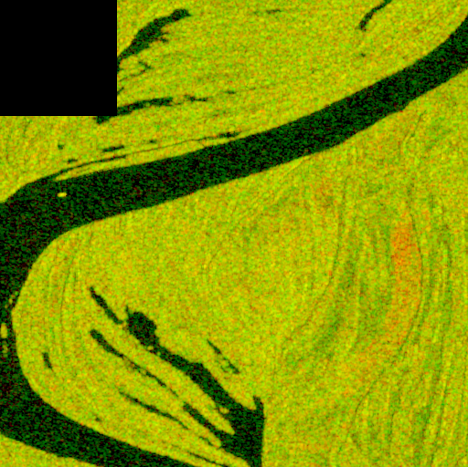}  
                                          & Replace the perturbed region with a constant value equal to the minimum pixel value of the corresponding image channel, computed from the channel's pixel distribution. \\
\textbf{Class mean}                       & \includegraphics[width=1.5cm]{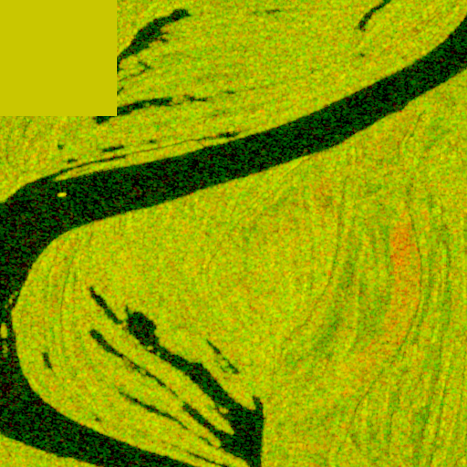}  
                                          & Replace the perturbed region with a constant value equal to the mean pixel value of a user-defined class of the corresponding image channel, computed from the channel's pixel distribution. \\
\textbf{Stochastic constant replacement}  & \includegraphics[width=1.5cm]{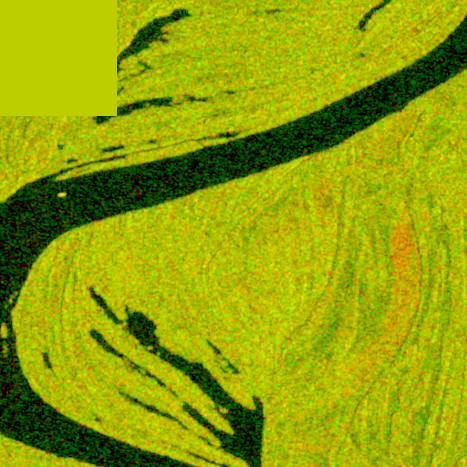}  
                                    	   & Sample a single constant value from a normal distribution fitted to pixel values in a local neighborhood around the perturbation region (i.e. a margin of size $w/2$), and use this value to replace the entire region.\\
\textbf{Patch substitution }              & \includegraphics[width=1.5cm]{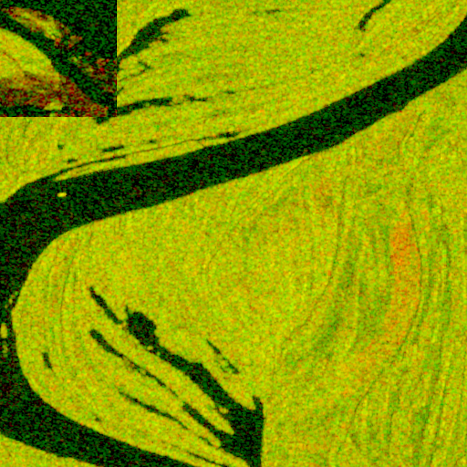}  
                                          & Replace the perturbed region with a randomly selected patch from a different sample in the dataset.\\
\textbf{Pixel shuffling}                  & \includegraphics[width=1.5cm]{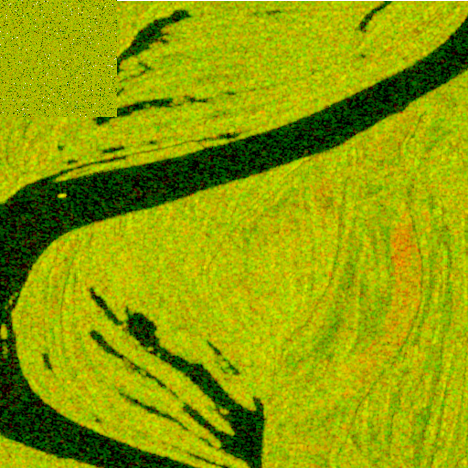}  
                                          & Randomly shuffle the pixels within the perturbed region, preserving the original pixel values but destroying their spatial arrangement.\\
\bottomrule
\end{tabularx}
\end{table*}

\emph{I. Estimating the Class Probabilities}\\
For an image $x \in \mathbb{R}^{H \times W \times C}$, where $H$, $W$, $C$ denote the height, width and channel dimensions, respectively, we create a set of $n$ locally perturbed versions $\{\hat{x}_i\}_{i=1}^{n} \in \mathbb{R}^{H \times W \times C}$ by systematically replacing a subset of the original pixel values with new values. 
Perturbations are defined by the geometry of the affected region, including its size and shape, whether or not they overlap, and by the perturbation type, which specifies the replacement value.
To investigate the effect of the patch geometry, we define the perturbation area either as a square window of fixed size $w$ or as an irregularly shaped superpixel that aligns more with the image structure. 
The overlap of perturbation regions is defined by the stride $s$ of their placement.
Note that this is only of relevance for square-shaped regions, as superpixels do not overlap. 
Placing a perturbation region over each pixel would lead to the spatially most precise relevance maps but is computational infeasible for common image sizes. 
Typical choices are a stride of either 50\% or 100\% of the size of the perturbation region, i.e. 50\% and no overlap, respectively.
For the perturbation type, we consider five replacement strategies, which are described in Table~\ref{tab:ptype}.
\textit{Channel-wise minimum} and \textit{class mean} are deterministic constant value baselines with a well-defined statistical meaning that are often the default choice in the corresponding literature \citep{occlusion, lime}.
In contrast to these two baselines, \textit{stochastic constant replacement} adapts to the local context around the perturbed region allowing to investigate the importance of spatial relationships and proximity in relevance estimation \citep{pred_differnce}.
\textit{Stochastic constant replacement}, \textit{patch substitution} \citep{bluecher_24} and \textit{pixel shuffling} are stochastic process.
Thus, we repeat the sampling process several times and average the resulting relevance maps to obtain a more robust estimate. 
The segmentation network $f_\theta$ then estimates the pixel-level class probabilities of the original image $y(x) \; = \: f_{\theta}(x) \in [0,1]^{H \times W \times K}$ and the set of perturbed images $\{{y}(\hat{x}_i)  \; = \: f_{\theta}(\hat{x}_i)\}_{i=1}^{n} \text{ with } {y}(\hat{x}_i) \in [0,1]^{H \times W \times K}$, where $K$ denotes the number of classes.

\emph{II. Computing the Probability Difference}\\
We calculate the difference between the class probability map of the original image $y(x)$ and each class probability map of the perturbed set of images $y(\hat{x}_i)$.
Subtraction is performed element-wise, i.e. $\Delta_i \; = \: y(x) \, - \, y(\hat{x}_i)$, resulting in a difference map for each pair of probability maps $\{\Delta_i\}_{i=1}^{n} \text{ with } \Delta_i \in[-1,1]^{H\times W\times K}$.

\emph{III. Defining the Attribution Area}\\
The attribution area is a user-provided input and consists of two parts: A binary segmentation map~$\alpha\in \{0, 1\}^{H\times W}$ of the same size as the input image, encoding pixels belonging to the attribution area by $1$ and remaining pixels by $0$, and a class ID~$\omega$. 
It formulates the underlying question "Which input regions influence the prediction of class~$\omega$ in region~$\alpha$?".
A natural choice for $\alpha$ is the predicted region of class $\omega$, i.e.
\begin{equation}
    \alpha(h,w) = \begin{cases}
        1 & \text{if } \arg\max y(h,w) = \omega, \\
        0 & \text{otherwise.} 
    \end{cases},
\end{equation}
where $y(h,w)$ is the predicted class posterior at pixel $(h,w)$.

From $\alpha$ we create attribution maps $\{\alpha_i\}_{i=1}^{n} \text{ with } \alpha_i \in \{0, 1\}^{H\times W}$, in which we systematically omit the perturbed area by setting it to $0$.
In this way, we avoid introducing a potential bias towards the perturbed areas in the relevance computation, which could arise from the larger probability changes within them.

\emph{IV. Estimating the Relevance Score}\\
We quantify the relevance of each perturbed input region with respect to the attribution area from the computed probability differences.
First, we extract the attribution-class channel from the difference maps, yielding $\{\Delta_i^\omega\}_{i=1}^{n} \text{ with } \Delta_i^\omega\in[-1,1]^{H\times W}$.
For each perturbed input region, we then multiply the respective difference map $\Delta_i^\omega$ element-wise with the attribution map $\alpha_i$ and average over the attribution area, i.e. the number of pixels in the attribution area $N_i$ according to 
\begin{equation}\label{eq:ri}
    r_i = \frac{1}{N_i}\sum_{h=1}^{H}\sum_{w=1}^{W}\Delta_i^\omega(h,w)\cdot\alpha_i(h,w),
\end{equation}
where
\begin{equation}\label{eq:ni}
    N_i = \sum_{h=1}^{H}\sum_{w=1}^{W} \alpha_i(h,w) .
\end{equation}

The resulting scalar $r_i \in \mathbb{R}$ is the relevance score associated with the perturbed area, i.e. how much perturbing this area influences the prediction of class~$\omega$ inside of the attribution area (while the perturbation area itself is excluded).
Repeating this for all $n$ perturbations, yields relevance scores $\{r_i\}_{i=1}^{n}$ for each perturbed input area.
Finally, we construct a dense relevance map $R \in \mathbb{R}^{H\times W}$. 
For square-shaped regions, we assign the score $r_i$ to the center pixel of the perturbed region and use either nearest-neighbor or linear interpolation to fill the gaps in between.
In the case of superpixel-based perturbation regions, the score $r_i$ is assigned to the complete corresponding perturbed superpixel.

\section{Experimental Setup}
Our study is grounded in the use case of flood detection from SAR imagery, outlined in Section~\ref{sec:usecase}.
The central component of our work is a systematic parameter study of the perturbation pipeline, for which we specify the parameter settings in Section~\ref{sec:occlusion}.
Complementing qualitative inspection, Section~\ref{sec:reldistri} describes the relevance distribution analysis that captures where relevance accumulates within the heatmaps.
Finally, Section~\ref{sec:eval} presents our evaluation approach to assess explanation quality, focusing on whether the relevance maps faithfully reflect the model's behavior.

\subsection{Use Case}\label{sec:usecase}
We formulate the detection of flooded areas in SAR images as a semantic segmentation task and employ standard a U-Net model \citep{Unet}. 
We train and evaluate it on a subset of the Sen1Floods11 \citep{Sen1Floods11} dataset, which comprises about 450 manually labeled Sentinel-1 images acquired in dual polarization (VV and VH), randomly split into training, validation, and test sets in a 60–20–20 ratio.
Covering eleven global flood events, the 512$\times$512 data chips compose a diverse dataset that includes large-scale floodplains, small water bodies, as well as potential confounding samples such as mountain shadows that appear similarly dark as water surfaces in SAR images.

\subsection{Parameter Study}\label{sec:occlusion}
Within the perturbation pipeline, we examine the two most critical parameters that affect the relevance score estimation:
The geometry of the perturbed region, including its size and shape, and the perturbation type, which specifies how pixel values inside the region are replaced.
To investigate the influence of the patch geometry, we define the perturbation area in two ways:
The first option is to create square patches and systematically shifting them across the image.
We use window sizes of $w = 8\times8$, $w = 32\times32$, and $w = 128\times128$ pixels for small, medium-sized, and large perturbation areas, respectively. 
With a stride of $s = w/2$ and $s = w$, we generate overlapping and non-overlapping patches.
As an alternative, we segment the image into superpixels via SLIC \citep{slic} using a segment size that matches the size of the square patches.
To examine the influence of the replacement value, we consider five perturbation types that progressively differ in how much of the original signal and structure they preserve (see Section~\ref{sec:method} Table~\ref{tab:ptype}).
One of the largest challenges in choosing an appropriate perturbation type is that perturbing an input region not only destroys the measured signal but inadvertently also creates a new signal. 
For example, the channel-minimum usually falls within the low backscatter range which is typically associated with water surfaces, making it a proxy for the water class.

\subsection{Relevance Distribution}\label{sec:reldistri}
\begin{figure}[h!]                  
  \centering
  \includegraphics[width=\linewidth]{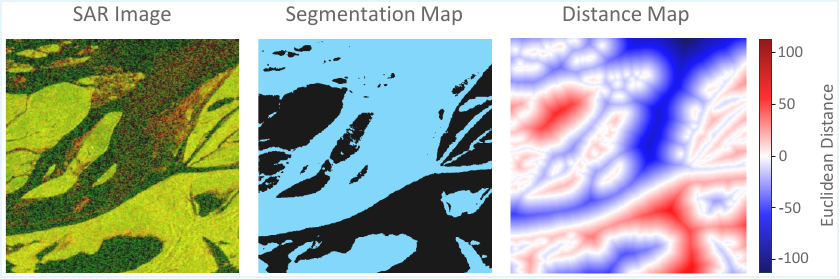}
  \caption{Example of a SAR image (false color composite) with the corresponding segmentation and distance map showing the Euclidean distance of every pixel to the water edge. Pixels on the water edge have a zero distance (white), water pixels have a negative distance (blue), and background pixels have a positive distance (red).}  
  \label{fig:dist_map}
\end{figure}

To assess where relevance concentrates within the estimated relevance maps and whether relevance patterns are consistent across perturbation strategies, we conduct a relevance distribution analysis. 
Specifically, we compute a signed distance transform of the segmentation map $\alpha$ using the water edge as reference, see Figure \ref{fig:dist_map}.
Each pixel is assigned its distance to the nearest edge pixel, with zero on the boundary, negative values inside water, and positive values in the background.
We then couple each pixel's relevance score with its signed distance value to form 2D histograms, where each bin counts pixels by joint distance-relevance values. 
These histograms are computed per image and averaged separately for each perturbation type using non-overlapping patches of size 32$\times$32 pixels.

Across perturbation strategies, the aggregated histograms visualize the distribution of relevance patterns relative to the water edge, and enable a direct, quantitative comparison, indicating whether relevance clusters within water, along the shoreline, or in the background.
Similar histograms indicate agreement about where class evidence is localized suggesting high consistency.
On the other hand, diverging histograms indicate low consistency implying relevance shifts.
When consistency is high, heatmaps converge on the same structures as being important for the prediction.
In such cases, what the network looks at appears independent of the perturbation approach, lending credence to the explanations. 
Low consistency, reflected by different histograms, suggests either instability or different internal reasoning across perturbation strategies.
In that case, explanations become method- or layer-dependent and should be interpreted carefully rather than cherry-picked.

\subsection{Correctness Evaluation}\label{sec:eval}
Beyond visual inspection, we assess explanation quality with respect to the correctness property~\citep{eval_nauta_23}. 
It measures how faithfully an explanation reflects the model's reasoning and indicates whether the highlighted regions correspond to what the model actually uses for its prediction.
Ideally, an explanation should reflect the model’s true behavior, and thus high correctness is desired.
To evaluate correctness, we adopt the deletion-based approach introduced in RISE~\citep{rise} and consider relevance maps produced across all perturbation types using non-overlapping patches of size 32$\times$32 pixels.

We progressively remove input regions according to their relevance map ranking and monitor the resulting change in the target class probability. 
A typical choice to remove input regions is to add Gaussian noise of a certain magnitude. 
However, initial experiments showed that this leads to the systematic effect that perturbed regions are consistently recognized as water, regardless of their original class, while true water regions were not reliably detected.
This shows that deletion-based evaluation is sensitive to the deletion scheme~\citep{bluecher_24}, just as perturbation-based explanations are sensitive to the perturbation design - and thus need to be used with care.
Thus, to properly simulate the absence of a class in our binary setting, where no additional class is available as an alternative target, we employ a class-conditional deletion scheme.
Specifically, water pixels are replaced by the channel-wise mean of the background pixels in the respective image, while background pixels are replaced by the channel-wise mean of water pixels in the respective image, ensuring the model confuses the correct class for the wrong one.

Starting with the most relevant input regions, we then remove regions with zero relevance, and finally negatively scored regions.
The underlying intuition is that if an explanation is correct, it identifies input regions on which the model truly relies.
Removing these regions therefore erases supportive evidence, leading to an early and pronounced drop in the predicted class probability.
We monitor the resulting change in the water class prediction relative to the reference data using a Dice score adapted to the fact that the reference labels contain a cloud class, whereas our task is defined over water and background only:

\begin{equation}\label{eq:dice_ours}
    Dice \; Score \;=\; \frac{2 \, I + 1} {\sum \, y(\hat{x}_i) \, \alpha \, (1 \, - \, g^c) \; + \; \sum \, g^w \, \alpha \; + \; 1} \;, 
\end{equation}

with

\begin{equation}
    I \;=\; {\sum \, y(\hat{x}_i) \, \alpha \, g^w} \;
\end{equation}

denoting the intersection term, corresponding to the true positive component in the Dice score, that is computed with respect to the reference water mask $g^w$.
The term $y(\hat{x}_i)$ denotes the prediction obtained from the perturbed input $\hat{x}_i$. 
The change in prediction is integrated over the attribution area $\alpha_i$.
We further consider a cloud mask $g^c$, computed from the reference label, to reduce the influence of cloud covered regions on the computation.
Additionally, we discard test images with high cloud coverage in the evaluation, reducing the test set to 44 instances.

We report the Dice score as a function of the fraction of removed regions, as well as the Area Under the Curve (AUC). 
A faithful explanation should yield a sharp initial drop when the most relevant regions (MRR) are removed, stabilize as increasingly irrelevant regions are perturbed, and may even increase when the perturbation eliminates disturbing inputs. 
This behavior corresponds to a medium area under the prediction curve.


\section{Results and Discussion}
In the results of our parameter study, we observe recurring patterns and structural similarities among individual relevance maps: 
For some data points, water is relevant for the prediction, while the background has a disturbing influence. 
For other instances, we notice the opposite pattern. 
Whereas in some relevance maps both classes equally either enhance or disturb the prediction, in others, only one class has a determining effect.
Beyond these patterns, we also notice a high variability in the relevance scores across different data points, even within the same parameter combination.
These observations indicate a formation of clusters and motivate the analysis of potential correlations between the relevance maps and the statistical properties of the images prior to discussing the results.

As a starting point of the cluster analysis, we chose the relevance maps resulting from the parameter combination with non-overlapping $32\times 32$ patches and the background mean as the replacement value, as we observe the strongest differences between the relevance maps with this setting. 
In total, we define seven properties describing the data points and relevance maps.
We formulate input data properties based on the segmentation map $\alpha$.
These properties describe the general composition of an input sample by the water extent, and the size and shape of water bodies. 

\begin{itemize}
    \item \textbf{Water coverage}: Ratio of water pixels $\sum_{h=1}^{H}\sum_{w=1}^{W} \alpha(h,w)$ to the total number of pixels $H\cdot W$.
    
    \item \textbf{Largest water component}: Size of the largest connected water component.
    We identify a set of $m$ water components $C_\omega(\alpha) = \{C_1, ..., C_m\}$ in the segmentation map $\alpha$, where each component $C_i$ is defined by connected water pixels. 
    We compute the maximal area over all components as $A = \max_i A(C_i) = \max_i |C_i|$.

    \item \textbf{Water area irregularity}: The irregularity of a water component $C_i$ is computed from its squared perimeter $P(C_i)^2$ over its area $A(C_i)$.
    This value is higher for river courses and lower for compact water structures.
    We then average the individual values over all water components. 
\end{itemize}

We further define properties describing the relevance maps based on the distribution of relevance scores.
By using the segmentation map $\alpha$ as a mask to assign relevance scores to each class, we compute the mean and standard deviation of relevance values for the water and background, $\bar{r}_{\mathrm{water}}$, $\sigma_{r,\mathrm{water}}$, $\bar{r}_{\mathrm{background}}$, $\sigma_{r,\mathrm{background}}$, respectively.

\begin{figure*}[ht]   
  \centering
  \includegraphics[width=\textwidth]{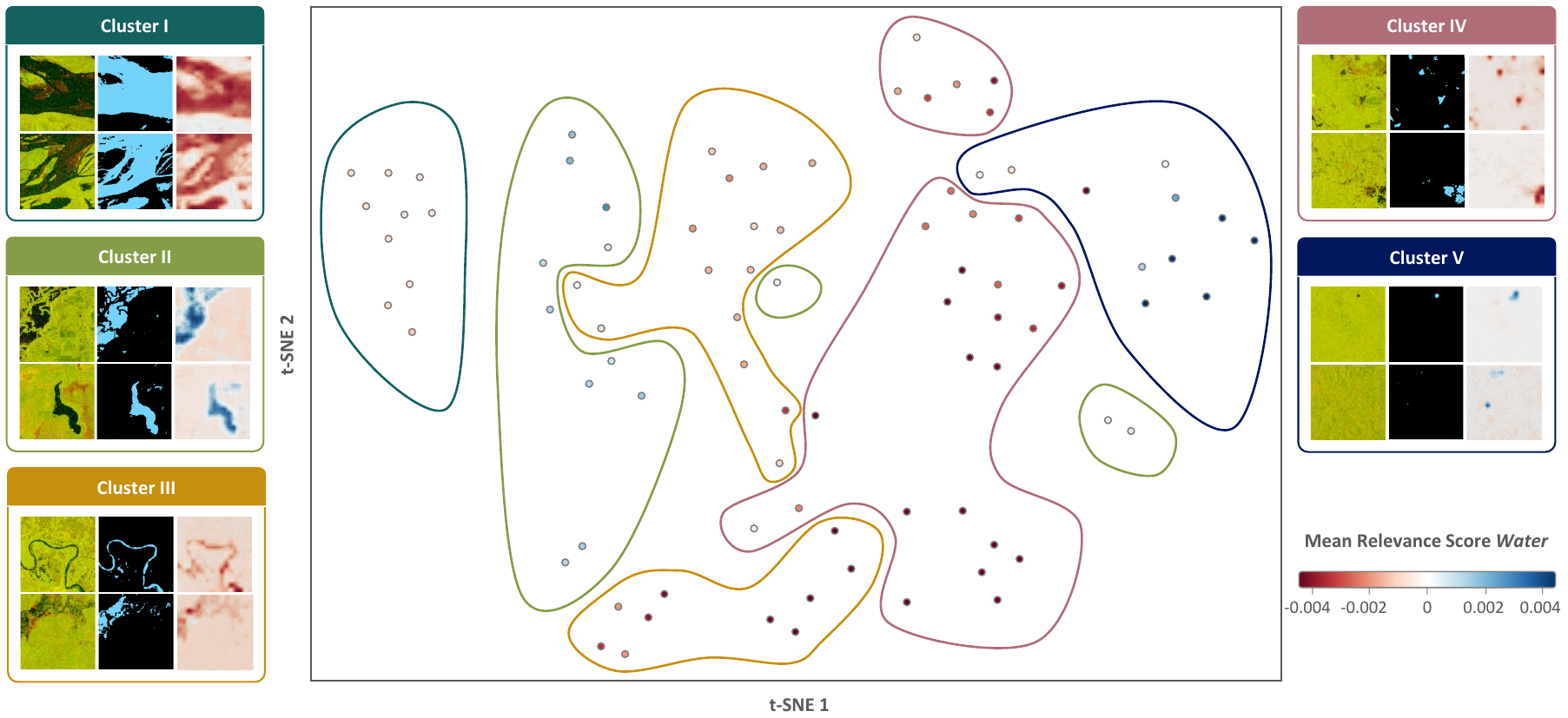}
  \caption{Overview of the cluster structure obtained from the t-SNE embedding of samples described by image and relevance map properties. Data points are colored according to the mean water relevance. Five clusters can be identified in the embedding space. For each cluster, two representative examples are shown, including the input image, the segmentation map, and relevance map used for the analysis.}
  \label{fig:cluster}
\end{figure*}

\begin{table*}[h!]
\centering
\small
\caption{Qualitative characterization of the five clusters identified in the t-SNE embedding. Each cluster is described by the predefined data and relevance map properties.}
\label{tab:clusters}
\renewcommand{\arraystretch}{1.5} 
\setlength{\tabcolsep}{5pt}
\newcolumntype{Y}{>{\centering\arraybackslash}X}
\begin{tabularx}{\linewidth}{Y Y Y Y Y Y}
\toprule

                                & \textbf{Cluster I} 
                                & \textbf{Cluster II }
                                & \textbf{Cluster III }
                                & \textbf{Cluster IV }
                                & \textbf{Cluster V} \\
\hline
\hline
\textbf{Water Coverage}         & extremely high  
                                & high 
                                & moderate 
                                & low 
                                & extremely low\\
\hline
\textbf{Water Body Composition} & large compact \& smooth or branched \& fragmented water bodies 
                                & mainly wide river courses \& larger water fragments that form complex shapes 
                                & mainly narrow river courses \& smaller water segments that form complex shapes
                                & mainly tiny river branches and scattered water fragments
                                & tiny individual water spots \\
\hline
\textbf{Relevance Scores}       & low
                                & medium range
                                & medium range
                                & high
                                & high \\
\hline

\textbf{Water Relevance}        & negative, small variance  
                                & positive, small variance
                                & negative, large variance
                                & negative, small variance
                                & positive small variance \\
                                
                                & water interferes with the prediction
                                & water enhances the prediction
                                & water interferes with the prediction
                                & water interferes with the prediction
                                & water enhances the prediction \\
                                
\hline
\textbf{Background Relevance}   & $\approx 0$ or positive, small variance
                                & negative, small variance
                                & $\approx 0$ or negative, large variance
                                & $\approx 0$ or negative, small variance
                                & $\approx 0$ or negative, small variance\\

                                & background is almost irrelevant or enhances the prediction
                                & background interferes with the prediction
                                & background is almost irrelevant or interferes with the prediction
                                & background is almost irrelevant or interferes with the prediction
                                & background is almost irrelevant or interferes with the prediction\\
\bottomrule

\end{tabularx}
\end{table*}

To visualize the similarities between the data points along the defined properties, we apply t-SNE~\citep{tsne}.
It creates a probability distribution representing these similarities between neighboring data points in the seven-dimensional space and maps them into a two-dimensional space.
Although t-SNE does not explicitly assign the data points to clusters, their spatial distribution in the projection in Figure~\ref{fig:cluster} shows multiple cluster formations of data points with similar properties. 
This suggests correlations between the statistical properties of the images and the estimated relevance scores, particularly the mean water relevance indicated by the coloring of the points.
Based on this t-SNE mapping, we identify five clusters.
Figure~\ref{fig:cluster} shows representative examples for each cluster, while Table~\ref{tab:clusters}  summarizes their characteristic properties.
Clusters differ with respect to both, the image characteristics and relevance patterns. 
With respect to the data properties, they vary in water coverage and water body composition.
Clusters~I and~II show a comparatively high water coverage with large coherent water bodies, whereas Clusters~III to~V are characterized by a low water coverage with small fragmented water structures. 
Further, clusters differ in the magnitude and distribution of relevance scores across water and background regions. 
In Clusters~II and~V, water regions tend to contribute positively to the prediction, while the background has a disturbing influence. 
Cluster~I shows the opposite pattern, while in Clusters~III and~IV, both classes have disturbing contributions.

\subsection{Effect of the Patch Geometry on the Relevance Estimation}
\begin{figure*}[h!]   
  \centering
  \includegraphics[width=0.91\textwidth]{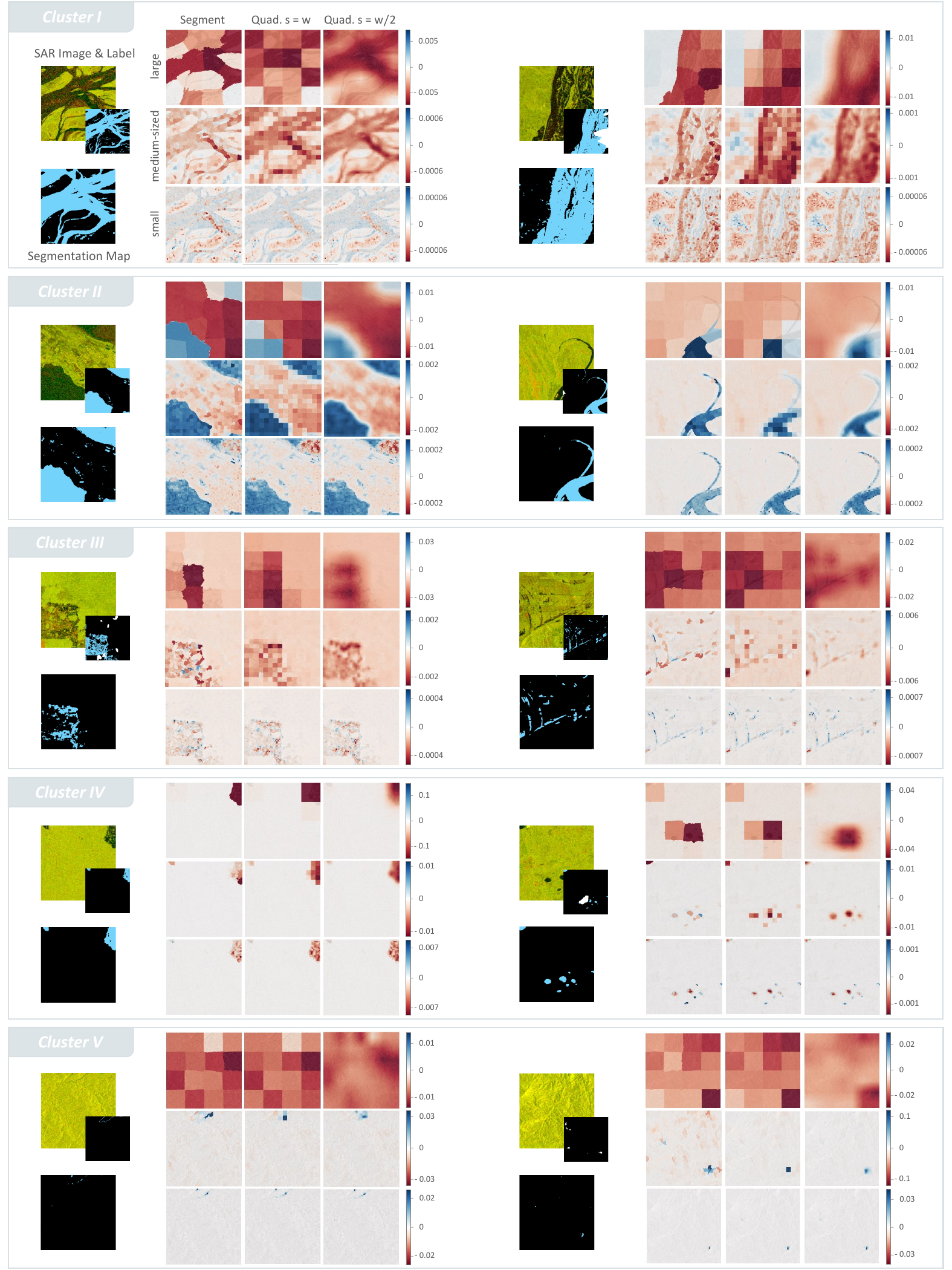}
  \caption{Influence of the patch geometry on the relevance estimation presented across the five clusters. For each cluster, two representative examples are shown, including the input image, label, segmentation map, and the corresponding relevance maps obtained for the class mean perturbation type. The relevance maps are compared across three patch sizes (large, medium, small) and three perturbation area definitions (superpixel segments, quadratic non-overlapping patches, where the stride equals the window size $s = w$, and quadratic overlapping patches, where $s = w/2$).}
  \label{fig:patch_size}
\end{figure*}

To investigate the influence of the patch geometry on the relevance estimation, we vary patch sizes and shapes while using the same perturbation type (class mean).
Among all perturbation types, the class mean, which is computed from the background pixel distribution, produces relevance maps that most clearly reveal differences induced by variations in the patch geometry. 
Figure \ref{fig:patch_size} presents an overview of the results for two samples of each cluster. 
A qualitative inspection of the results suggests three key aspects in which the choice of the patch geometry affects the relevance estimation.

\emph{1. Magnitude of the Relevance Scores}
We observe that the relevance scores fall within the same value range for different patch shapes of similar size. 
However, when comparing the relevance scores across different patch sizes, the value range varies and the relevance scores decrease as the patch size decreases.
Since the relevance of a patch is based on the change in the model's prediction, altering a larger portion of the input yields larger prediction shifts, which in turn reflects in a higher relevance estimation.
This relationship becomes clear in our experiments: 
The largest patches have a size of $128\times 128$ pixels and thus cover 1/16 of the input image.
Compared to the smallest patches, which cover an area of about $8\times 8$ pixels, their influence on the model's prediction and relevance estimation is significantly larger.

\emph{2. Aggregation of Contributions}
Varying the size and shape of the perturbation patch can lead to different relevance estimations for the same input region. 
For instance, perturbing a narrow river branch with a large square patch will yield a different relevance estimate for the river branch compared to using a superpixel-based segment that closely follows the river's contours. 
This effect is observable particularly in the samples of Cluster~I and~V.
Relevance scores computed for large perturbation patches therefore represent an aggregate of contributions from multiple subregions within the perturbed area (mixed patches).
Due to this aggregation, it is impossible to disentangle the individual contributions and to determine precisely which part of the perturbed area influenced the relevance estimation and in what way.
In contrast, smaller perturbation patches target more localized structures and features in the input, allowing for a more direct attribution of the relevance score.
Superpixel-based segments that follow the boundaries of the underlying image structures tend to reduce this aggregation effect, especially when the segment size closely aligns with the image features.

\emph{3. Resolution of the Relevance Maps}
Lastly, the size and shape of the perturbation patch play a central role in determining the resolution and thus the interpretability of the relevance maps. 
A comparison between small and large patches reveals that larger perturbation areas typically produce coarse and fragmented relevance maps, which do not accurately reflect the underlying image structures. 
Such relevance maps tend to blur fine-grained details and obscure class-specific information, thereby reducing their interpretability. 
In contrast, smaller perturbation patches yield relevance maps of higher resolution. 
These relevance maps are more sensitive to local image features, better differentiate between semantic classes, and reveal the structures that contribute to the model’s prediction more precisely.
However, using small perturbation patches can significantly increase the run time.
Beyond the size of the perturbed area, it's shape also significantly influences the resolution of the relevance maps.
Especially non-overlapping square patches often disrupt the spatial continuity of object boundaries or semantic regions.
This is particularly problematic with large patches, which fully replace smaller structures.
While overlapping patches mitigate this issue to some extent, superpixel segments prove to capture the underlying image information best.
Since such segments are derived from the inherent image structure, they better preserve coherent structures and semantic information when being perturbed, resulting in relevance maps that align with the underlying image information more closely. 

\emph{4. Runtime}
With an increasing number of perturbation patches to be evaluated, runtime increases. 
For quadratic perturbation patches, runtime is determined by both, the patch size $w$ and stride $s$.
In the non-overlapping case, runtime scales approximately with $1/w^2$ with varying patch sizes, while for a fixed patch size and varying stride it scales approximately with $1/s^2$.
For segment-based perturbations, runtime is primarily determined by the number of segments.
Smaller segments increases runtime due to the larger number of regions to be evaluated.

\emph{Summary} The patch geometry used in perturbation-based relevance estimation has a substantial impact on the resulting relevance maps and their interpretability. 
Despite their overall larger influence on the prediction, large perturbation areas, particularly non-overlapping squares, tend to produce coarse-grained relevance maps. 
Their low spatial resolution obscures finer image details, making it difficult to attribute relevance to specific image structures, and thereby offering limited insight into the model's decision-making process.
In contrast, smaller perturbation patches or shapes that are better aligned with the image structures produce high-resolution relevance maps that reflect the spatial and semantic composition of the image more accurately. 
While the generally low relevance scores indicate a weaker influence on the prediction, the relevance scores can be assigned to meaningful structures more precisely, which enhances the understanding of the model's predictions.
Overall, the analysis reveals a trade-off between the influence a patch has on the prediction and the detail with which relevance can be spatially localized. 
Considering that perturbation-based relevance estimation aims to provide  insights into the model's decision-making process, the choice of patch geometry presents a crucial parameter that directly impacts the interpretability of the relevance maps and thus the reliability of the explanations.

\subsection{Effect of the Perturbation Type on the Relevance Estimation}
\begin{figure*}[h!]   
  \centering
  \includegraphics[width=0.91\textwidth, trim={0cm 0.1cm 0cm 0.1cm}, clip]{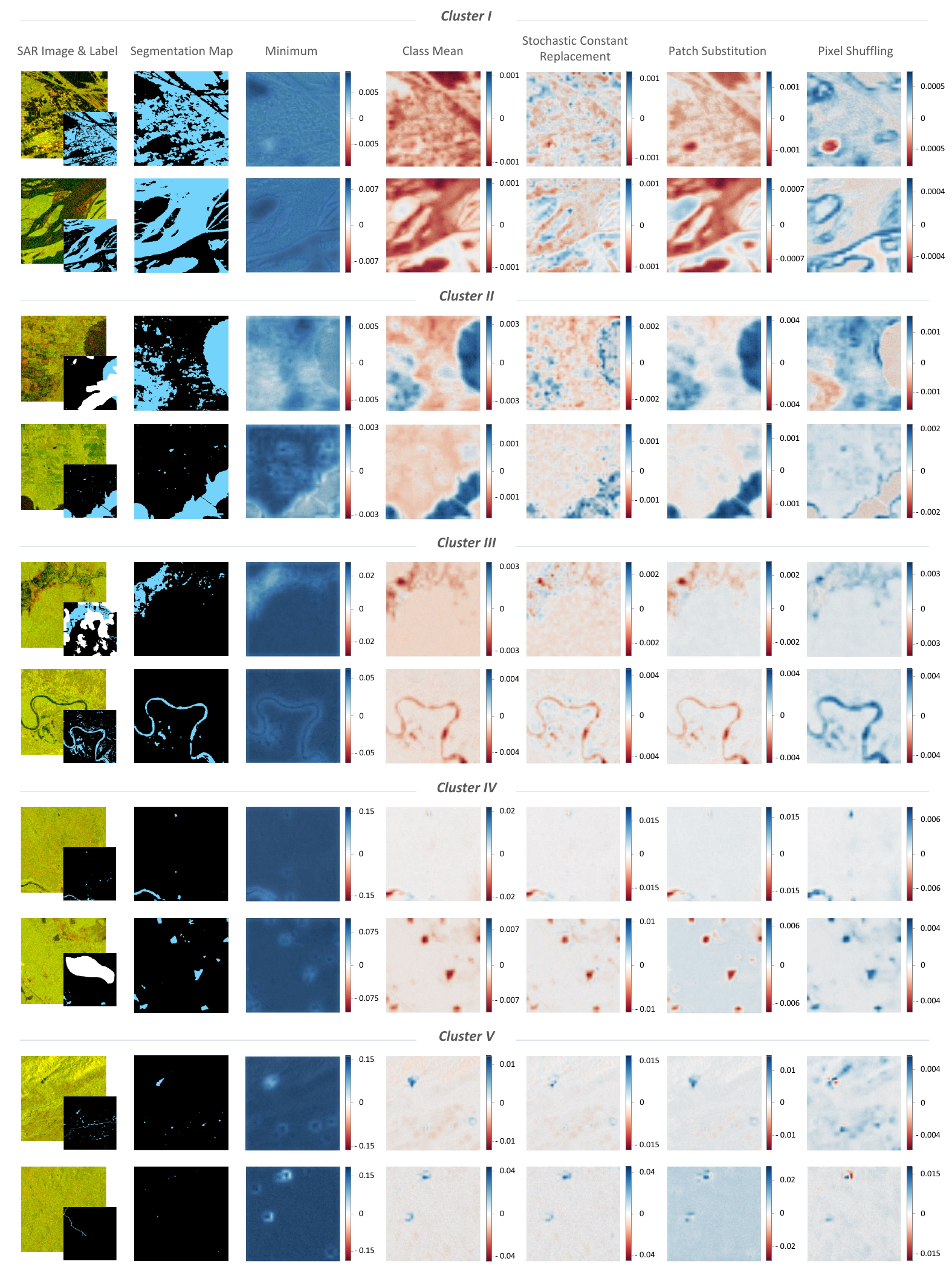}
  \caption{Influence of the perturbation type on the relevance estimation presented across the five clusters. For each cluster, two representative examples are shown, including the input image, label, segmentation map, and the corresponding relevance maps obtained with overlapping $32 \times 32$ patches ($s = w/2$). The relevance maps are compared across five perturbation types, including minimum, class mean, stochastic constant replacement, patch substitution, and pixel shuffling.}
  \label{fig:occ_value}
\end{figure*}

\begin{figure*}[h!]   
  \centering
  \includegraphics[width=\textwidth]{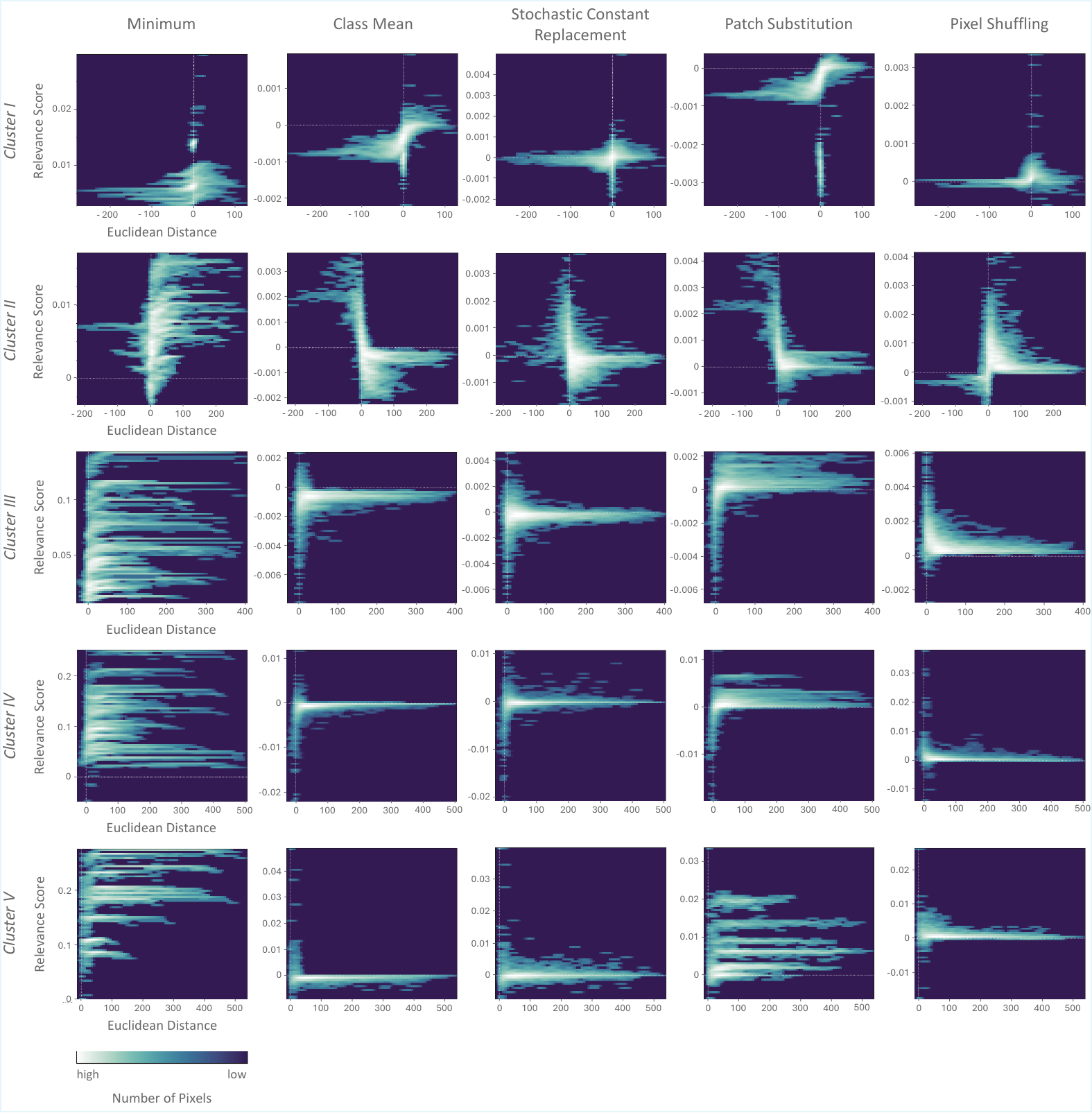}
  \caption{Average relevance distribution across clusters and perturbation types for overlapping $32 \times 32$ patches ($s = w/2$). For each cluster (row) and perturbation type (column), we show a 2D histogram of relevance scores as a function of Euclidean distance, averaged over all samples in the respective cluster. Color intensity reflects the number of pixels contributing to each bin, with brighter regions indicating higher pixel density.}
  \label{fig:dist_transform}
\end{figure*}

To investigate the influence of the perturbation type on the relevance estimation, we compute relevance maps using overlapping $32\times 32$ patches ($s = w/2$). 
The chosen patch geometry offers a good balance between the spatial resolution to trace the underlying image structures and the impact on the prediction.
We present the results for all five perturbation types - the channel minimum, class mean, stochastic constant replacement, patch substitution, and pixel shuffling - across the identified clusters in Figures~\ref{fig:occ_value} and~\ref{fig:dist_transform}. 
Despite using the same window size, we obtain different relevance scores and distributions depending on how the image is perturbed.
While there are minor inter-cluster variations, each perturbation strategy exhibits a characteristic pattern that recurs across clusters. 

\emph{Channel minimum}
Among all perturbations, the minimum value yields the most distinct relevance maps with large, mostly positive scores for both water and background regions across all clusters (Figure~\ref{fig:occ_value}).
Background scores are generally higher, relevance is nearly constant within a sample and varies across samples.
This is reflected particularly in the histograms for Clusters~III-V in Figure~\ref{fig:dist_transform}.
For the dominating background class relevance is distributed as multiple narrow bands along the distance axis.
Water scores, in contrast, are generally lower and often taper toward zero. 
The water edge region is characterized by mixed patches where perturbation windows cover both classes.
Correspondingly, the histograms show dense concentrations of near-edge background pixels at lower relevance levels.
Such mixed patches are most prevalent in clusters with small, fragmented water bodies or narrow river courses, where contiguous water areas are smaller than the perturbation window, such as in Clusters~III-V (Figure~\ref{fig:dist_transform}).
This pattern recurs across clusters, indicating a stable response to the minimum value perturbation.
That stability likely follows from the minimum being an extremely low intensity which is less sensitive to the local content it overwrites.
At the same time, the band-shaped background distribution along the relevance axis also highlights strong intra-cluster variability despite the consistent overall trend, see especially Clusters~III-V in Figure~\ref{fig:dist_transform}.

Overall, the relevance maps suggest that primarily background regions contribute to the prediction.
Although the minimum resembles low water backscatter, it is an extreme value that poorly represents the variability of water returns and therefore suppresses informative water cues.
Yet it still tends to fall within what the model associates with water, as perturbing background pixels with the minimum yields a water response.
The inserted patch acts as an extreme, water-like outlier that stands in contrast against the surrounding background, replacing useful background evidence.
This harms the prediction and marks the perturbed region as important.

This analysis points to two interacting effects underlying the minimum value perturbation. 
First, it suppresses mixed scattering within water regions by injecting an extreme low signal that fails to capture class variability and cancels out informative water cues the model relies on.
Second, it amplifies the contrast between the extremely low value of the water-like patch and the higher-valued background, again acting as an outlier that confuses the model.
Together, these effects define a characteristic signature, with relevance maps dominated by strong positive scores across all clusters.

\begin{figure*}[h!]   
  \centering
  \includegraphics[width=0.61\textwidth]{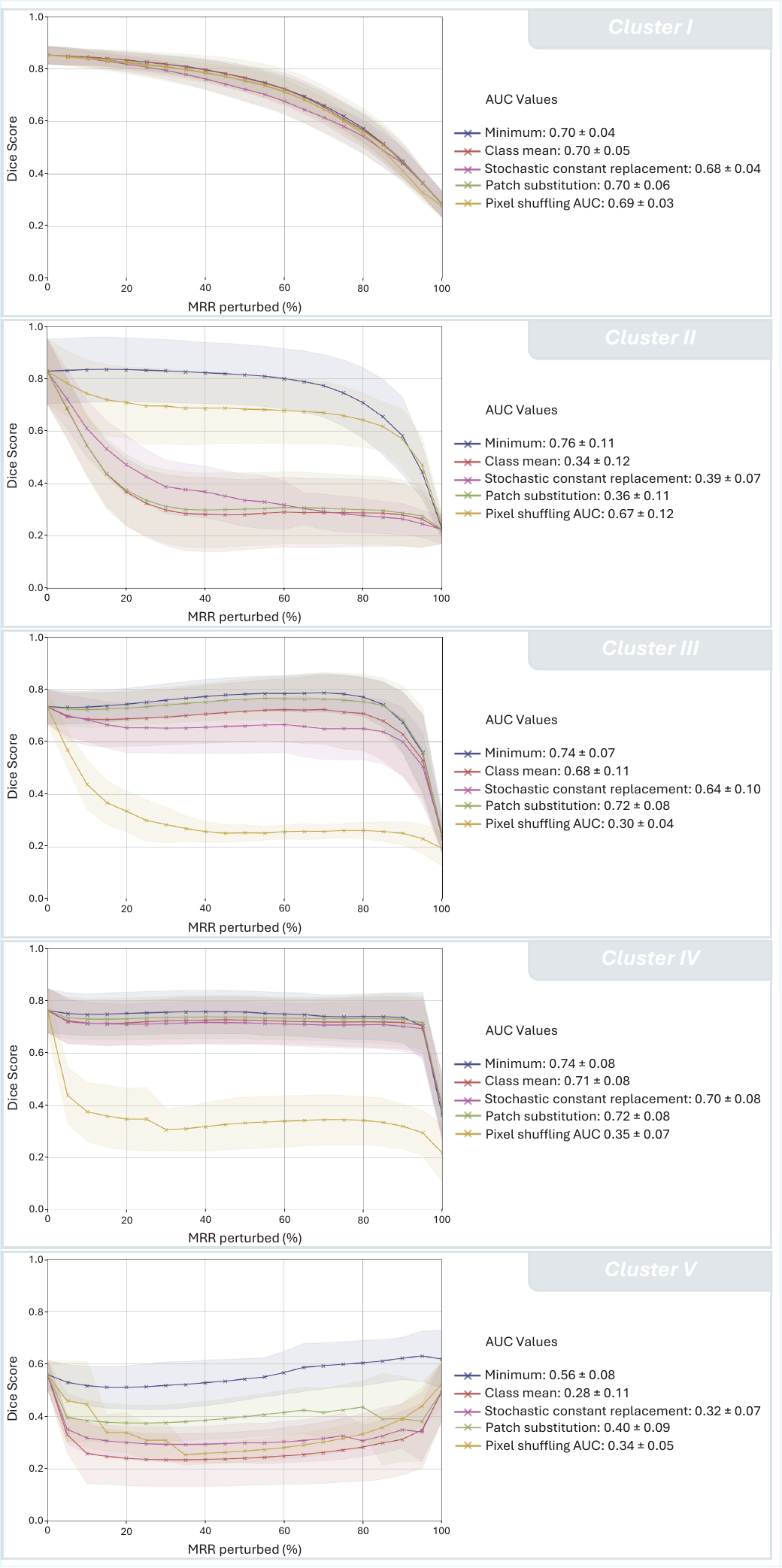}
  \caption{Correctness evaluation across clusters for different perturbation types using overlapping $32 \times 32$ patches ($s = w/2$). For each cluster, the curves show the average change in Dice score as the most relevant regions (MRR) identified by the respective relevance maps are progressively perturbed. Colors indicate the five perturbation types, and the reported AUC values summarize the overall degradation behavior.}
  \label{fig:eval_curves}
\end{figure*}

To complement the qualitative analysis, Figure~\ref{fig:eval_curves} summarizes the faithfulness evaluation based on deletion, tracing how the prediction changes as input regions estimated to be relevant are progressively removed.
Across clusters, the minimum value curves remain comparatively shallow and yield consistently high AUCs, typically above 0.7.
Early in the deletion, the Dice score stays relatively stable close to 0.8 and declines only slowly, despite perturbing regions ranked as most relevant.
Toward the end, the Dice score drops sharply, in some cases down to 0.2, once regions with lower relevance or counter evidence are removed.
In the minimum value heatmaps in Figure~\ref{fig:occ_value}, background is consistently ranked as more relevant than water, so the deletion procedure perturbs background first.
The flat early behavior suggests that the top ranked regions in these heatmaps are not decision critical for the model, or that minimum value replacements introduce artifacts that the model largely discounts during deletion.
This aligns with the earlier observation that the minimum acts as an outlier in both water and background scenes and is comparatively insensitive to the signal it overwrites.
The Dice score then degrades once deletion reaches water regions. 
Notably, the point where this degradation becomes apparent shifts with the water coverage across clusters.
In clusters with large water extent, the drop starts earlier, when roughly 60\% of the most relevant input regions are removed in Cluster~I and about 70\% in Cluster~II.
In increasingly background dominated clusters, the decline is delayed until a much larger fraction has been deleted, often only after most high relevance background has already been removed, around 85-95\% in Cluster~III and~IV.
Overall, the deletion curves indicate limited faithfulness for the minimum value perturbation, since removing its top ranked regions does not reliably break the prediction and the induced ranking does not align with the model’s most causal evidence as reflected by the delayed Dice degradation.

\emph{Class mean, Stochastic constant replacement, and Patch substitution}
Similar to the minimum value perturbation, the class mean perturbation replaces patches with a constant substitute but in this case computed from the background pixel distribution.
On background, the perturbation patch averages heterogeneous scatterings into a uniform surface, and on water, it recasts the water signature as background. 
However, unlike the minimum, the class mean is not an extreme value, but falls near the center of the background class distribution.
In background-dominated samples, the replacement remains close to the surrounding statistics, making it more sensitive to the underlying scene. 
On water, it can create a strong contrast to the low backscatter signal.
This effect becomes apparent in the relevance maps and histograms, which systematically differ across clusters, and in some cases show an inverted relevance pattern between water and background regions (see Figures~\ref{fig:occ_value}~and~\ref{fig:dist_transform}).

Water-heavy clusters (Clusters~I and~II) exhibit the clearest sign reversal.
In Cluster~I, water predominantly acts as counter-evidence while the background is rather irrelevant, whereas in Cluster~II, water becomes supportive and the background turns disturbing (see Figure~\ref{fig:occ_value}).
This inversion is not only reflected in the heatmaps but also captured in the histograms, which reveal complementary S- and Z-shaped relevance distributions (see Figure~\ref{fig:dist_transform}).
In Cluster~I, the histogram traces an S-like trajectory.
Background pixels concentrate near zero relevance along the distance axis while water pixels carry negative relevance scores that increase as moving deeper into the water interior.
Cluster~II mirrors the relevance distribution as a Z-like profile.
Here, water pixels have positive relevance that rises toward the water interior, while background pixels exhibit negative relevance that grows with increasing distance from the water edge. 

Background-heavy clusters, in contrast, are driven by water-focused response, as water regions stand out sharply against a neutral background.
The water response can be positive (Cluster~V) or negative (Clusters~III and~IV), but across all three clusters the background remains largely inactive. 
This class separation becomes increasingly pronounced as water coverage decreases from Cluster~III to Clusters~IV and~V. 
This is reflected in the histograms, which show similar distributions of near-zero background relevance along the distance axis.

When forming a prediction, the model reacts to cues, such as brightness, texture or contrast. 
The class mean perturbation indicates, that the model relies on different cues across the clusters and reweighs their importance from scene to scene.
Replacing counter-evidence of water, such as bright or rough spots within larger water areas, by a clean background patch can increase the contrast between the classes and thus the confidence in water (Cluster~I).
Removing supportive cues like clean dark water patches, on the other hand, can lower it (Cluster~II).
While smoothing textured background parts can resolve confusion in cluttered areas, lifting the water score, reducing the contrast and textual cues the model uses against the water class, can yield a drop in the water score. 
Along the water edge, replacing a patch with the class mean can suppress mixed signals, yielding a clearer class transition (Cluster~V).
At the same time, smoothing clear shorelines can reduce useful water cues, hurting the prediction (Clusters~III and~IV).

We observe similar relevance patterns in both heatmaps and histograms when patches are replaced by values sampled from the local pixel distribution (stochastic constant replacement) or by patches extracted from other images in the dataset (patch substitution), see Figures~\ref{fig:occ_value}~and~\ref{fig:dist_transform}.
Compared with the class mean perturbation, where a patch is replaced by a constant value, stochastic constant replacement draws values from the vicinity of each patch, yielding relevance maps with a higher local variability (see Figure~\ref{fig:occ_value}) and correspondingly more diffuse relevance distributions (see Figure~\ref{fig:dist_transform}).
Patch substitution, by contrast, tend to produce smoother relevance maps with stronger class contrast (see Figure~\ref{fig:occ_value}). 
For Clusters~I and~II, the patch substitution histograms show a stronger separation between water and background, whereas for Clusters~III-V they hint towards high intra-cluster variability, underscoring how strongly relevance responds to scene-specific statistics under perturbation. 
As patch substitution introduces entirely new content into the host image, they induce the strongest distribution shift among all perturbation types.
Since both perturbation strategies are stochastic, we compute relevance maps multiple times and average them to obtain a more robust estimate. 
Notably, the averaged relevance maps from these perturbation types converge toward class mean maps.

This convergence implies that the background class dominates the pixel distribution within individual images and across the dataset. 
Consequently, most random draws from a patch's local distribution, as well as most randomly selected patches from other images, resemble background.
On average, this reproduces the class mean case.
In contrast, when water coverage is high and the water class locally dominates or when classes are stronger intermixed (Clusters~I and ~II), stochastic constant replacement diverges most from the class mean baseline, yielding relevance maps with a higher local variability and thus more diffuse relevance distributions.
Background-heavy samples (Clusters~III-V), on the other hand, yield smoother relevance maps that resemble the class mean maps, along with relevance distributions that mirror those of the class mean strategy.
This suggests that the model is more sensitive to local statistics than to the scene-level appearance.
For patch substitution perturbations, deviations from the class mean case likely reflect cross-image differences in image statistics.
Although the dataset is normalized to a shared scale to stabilize training, image statistics are not identical, and neither are the cues the model relies on when forming predictions. 
The same class can therefore appear smoother or rougher in the substituted patch compared to the host image, introducing stronger shifts in the relevance estimation and yielding clearer relevance differences between classes.
This, in turn, suggests that the model has learned the underlying class concepts, identifying supportive and suppressive cues across scenes. 

Taken together, the three perturbation types act as background surrogates in different ways.
The class mean provides a clean, local background replacement.
Stochastic constant replacement is a stochastic version of the same idea that approaches the class mean case as background prevalence grows.
Patch substitution adds out-of-image information, which, in a background-dominated dataset, represent background stand-ins at scale.
While the class mean probes the model's dependence on background texture versus smooth regions, stochastic constant replacement tests whether those dependencies are local. 
Patch substitution further reveals the model's sensitivity to out-of-image shifts. 
Both stochastic constant replacement and patch substitution expose the dataset's background dominance and thus reflect on the underlying class imbalance

Accordingly, all three perturbation strategies follow the same core concept, yielding relevance maps that share the broad structure and sign patterns.  
What the model counts as supporting or suppressing a prediction, depends on the scene composition, e.g. water roughness, background heterogeneity, and boundary density. 
These perturbation types reflect the model's sensitivity to within-class variability, which recurs in samples with similar image characteristics.
Their similarities arise from how closely each strategy matches the local statistics and how much spatial structure it erases, so the resulting explanations strongly depend on which signal is overwritten.

This dependency is also reflected in the faithfulness evaluation.
Across clusters, the deletion curves for stochastic constant replacement and patch substitution largely track the class mean curves, typically with slightly less extreme changes (see Figure~\ref{fig:eval_curves}).
Supporting the agreement already seen in the heatmaps and histograms, the evaluation suggests comparable faithfulness across the three strategies.
At the same time, the evaluation curves remain strongly cluster dependent. 
In clusters where background is prioritized over water (Clusters~I,~III, and~IV) deletion initially perturbs background regions.
The Dice score decreases only mildly and drops stronger once water regions are perturbed.
This curve behavior is consistent with water being decisive for the segmentation despite its lower ranked relevance in the heatmaps. 
The resulting AUC values are comparatively high, around 0.6 to 0.7, and resemble the behavior observed for the minimum value perturbation. 
In clusters where water is ranked as the dominant contributor (Clusters~II and~V) the Dice score falls steeply in the beginning.
In Cluster~II it reaches about 0.3 within the first 30\% of deletion and stabilizes afterwards. 
In Cluster~V the Dice score drops even faster, reaching a similarly low score within the first 10\%, followed by an extended plateau.
Notably, Cluster~V shows a late increase in the Dice score toward the end of deletion, yielding a parabola-shaped curve that matches the signed interpretation of the heatmaps.
Once supportive regions are removed, the curve plateaus while irrelevant regions are perturbed, and then rises once counter-evidence is removed.
We do not observe this late recovery in Cluster~II, which is consistent with its more balanced class composition.
Here, supportive evidence is distributed across broader regions and is therefore removed more gradually, so the curve reaches its plateau only after the prediction has already degraded substantially.
Once the Dice score has settled at a low level, removing remaining counter-evidence barely yields an improvement.

Overall, the evaluation reinforces the qualitative observations that class mean, stochastic constant replacement, and patch substitution perturbations behave similarly, while their faithfulness depends on scene composition and on the class the perturbation primarily targets. 
When water is prioritized, the early drop in the deletion curves confirms that the heatmaps capture evidence the model truly relies on. 
When background is prioritized, the shallow early decay and delayed drop indicate that the heatmaps actually miss prediction driving cues, underscoring that relevance estimates are shaped by the perturbation choice.

\emph{Pixel Shuffling} 
The pixel shuffling perturbation strategy yields characteristic relevance maps across all clusters, consistently highlighting boundaries that separate water from background rather than class interiors (see Figure~\ref{fig:occ_value}).
Along the water-background edge, the predicted probability drops and the relevance score peaks.
Relevance concentrates where preserving spatial coherence is essential.
By permuting pixels within the perturbation patch, pixel shuffling preserves local intensity statistics while disrupting spatial arrangement.
The boundary-focused relevance maps underscore the need for separable, contiguous structures in the model's decision-making.
Thus, pixel shuffling provides a complementary perspective on what the model relies on when forming its prediction.

In large, homogeneous regions, such as extensive water bodies (Clusters~I and~II) or broad background regions (Clusters~IV and~V), relevance is typically modest and spatially uniform (see Figure~\ref{fig:occ_value}).
Permuting pixels in such areas perturbs texture, but leaves class evidence largely intact. 
Over water, the prediction benefits from this, while it deteriorates over background, depending on which cues the model picks upon and whether disrupting those cues proves supportive or suppressive.
By contrast, structures smaller than the perturbation patch stand out in the relevance maps.
Narrow rivers (Cluster~III) and fragmented water structures (Clusters~IV and~V) are fully covered by the perturbation patch. 
Their structure is entirely disrupted, as pixels from both classes intermix.
The original structure is no longer recognizable as the correct class, and the prediction suffers accordingly.
Similar to the behavior over water edges, class mixing blurs the boundary and degrades separability between classes and individual structures.

This emphasis on class separation is also evident in the histograms in Figure~\ref{fig:dist_transform}. 
In Cluster~II, where water and background are comparatively balanced, the sign split is particularly clear, with water pixels occupying only the negative relevance quadrant while background pixels occupy only the positive relevance quadrant, leaving the remaining quadrants empty.
The histogram captures a class-consistent sign structure, with each class confined to its respective relevance sign.
This behavior shifts in Clusters~III through~V as samples become increasingly background-dominated and water appears more often as fragmented patches or narrow rivers, increasing the share of mixed patches.
Here, pixel shuffling fully destroys the signature of water, strongly harming the prediction. 
Accordingly, the histograms shift toward predominantly positive relevance with a sharp cutoff around zero. 
Unlike the clean two-quadrant separation in Cluster~II, Clusters~III-V concentrate almost entirely on the positive side of the relevance axis, reflecting the prediction degrading effect pixel shuffling has in fine-structure mixed areas.

Taken together, these patterns indicate that pixel shuffling probes the model's dependence on spatial arrangement and edge continuity rather than on pixel-level statistics alone.
In homogeneous regions, permuting pixels largely preserves class evidence, whereas in mixed areas it impairs the separation between water and background, hurting the prediction. 
The incremental deletion curves (Figure~\ref{fig:eval_curves}) mirror this distinctive behavior across clusters.
At the same time, the curves follow the same general mechanism observed for other perturbation strategies. 
When the heatmaps prioritize background over water (Clusters~I and~II), the Dice score falls only mildly at first and drops more strongly once water regions are perturbed, yielding comparatively flat curves with high AUC values near 0.7.
When the heatmaps prioritize water (Clusters~III-V), the Dice score drops sharply early in the deletion process and settles low, yielding smaller AUC values near 0.3.
For pixel shuffling, however, this pattern should be interpreted through its boundary-centered mechanism.
In Clusters~III through~V, fragmented water regions contain a large fraction of boundaries and are turned into mixed patches by pixel shuffling.
In the heatmaps, these regions appear highly relevant, which drives the early Dice collapse.
As a result, the curve behavior can give the impression that the heatmaps identify water as the decisive cue, even though the underlying effect is primarily the disruption of class boundaries.
Overall, pixel shuffling provides a robust and consistent edge sensitivity, revealing that the model does not rely on pixel statistics alone when forming a prediction, but strongly depends on continuous class boundaries.

\subsection{Summarizing the Effects on the Relevance Estimation} 
This analysis shows that relevance maps derived under different perturbation types reflect how that choice shapes the model's decision-making.
When forming a prediction, the model relies to varying degrees on different cues within each data point. 
Perturbation strategies shift these cues, for instance by altering texture, brightness, or spatial composition, and thereby regulate the extent to which these cues support or hurt the prediction. 
In other words, different perturbation strategies can amplify or attenuate these cues, which is reflected in the relevance maps as beneficial or distracting contributions to a prediction. 
Consequently, different perturbations yield different relevance maps and therefore different explanations. 
Our results suggest that the choice of the \textit{perturbation type} parameter strongly affects relevance estimation, calling into question the interpretability of perturbation-based xAI explanations.
The class-conditional deletion-based evaluation complements the qualitative inspection of the heatmaps by testing whether the relevance ranking produced by each perturbation strategy aligns with the model’s decision critical content, providing an estimate of faithfulness.
While this evaluation does not establish a single perturbation strategy as generally more faithful, it consistently shows that explanations prioritizing water over the background align better with the model’s prediction logic.

The minimum value replacement injects an extreme signal to the input, that creates a strong contrast to its surrounding.
Acting like an outlier, it overshadows scene-specific cues and substantially determines the relevance estimation. 
Across scenes, relevance concentrates on background rather than on water regions, that the model relies on, yielding explanations that deviate from the model’s internal reasoning.
In contrast, the class mean, stochastic constant replacement, and patch substitution strategies are more sensitive to the content they overwrite. 
Under these perturbations, the model leans more strongly on scene cues such as texture, edges, and contrast, as well as on class balance, and label quality.
Thereby, the class mean presents a smooth substitute that primarily probes the model's dependence on these cues, whereas the stochastic constant replacement strategy tests whether those dependencies are local.
Patch substitution, in turn, exposes sensitivity to out-of-image information.
Because these perturbation strategies interact with scene properties, relevance can concentrate in different regions across samples and can even invert between classes, so that faithfulness varies with the scene.
The similarity of the relevance maps produced by these three perturbation strategies suggests that they converge toward background stand-ins, which is consistent with the background dominated class composition.
Pixel shuffling preserves intensity statistics but disrupts structure, underscoring the importance of class boundaries and spatial arrangement. 
Although the deletion curves suggest that pixel shuffling can yield relevance rankings that appear faithful, this agreement may be coincidental. 
In scenes with small water fragments, pixel shuffling can break up the water pattern, creating mixed patches that are highlighted in the relevance maps as decision driving.
This can make water look relevant in the heatmap even though the model is mainly reacting to boundary disruption, which puts the faithfulness of these explanations into question.
It also indicates a limitation of the evaluation method, which can be insensitive to cases where perturbation assigns relevance that agrees with the model outcome for the wrong reason.

Taken together, these results show that some perturbation types behave comparatively robust across scenes (minimum, pixel shuffling), whereas others are more scene-sensitive and cluster with similar image characteristics (class mean, stochastic constant replacement, patch substitution). 
Overall, the perturbation type determines the relevance estimation and thus the ranking of input regions.
Thereby, scene-sensitive perturbations can provide relevance estimates that are more faithful to the model’s underlying decision logic.
Together with the observed effect of patch size on the relevance estimation, these findings reinforce that perturbation-based explanations are parameter dependent.
As the model's reliance on specific cues shifts with the chosen perturbation, the resulting relevance maps turn out ambiguous. 
Without explicitly accounting for how the perturbation alters the model's decision-making, the explanatory value of perturbation-based relevance maps remains uncertain. 
This undermines the reliability of the method and ultimately its ability to foster trust in the model’s predictions.

\section{Concluding Remarks}
Perturbation-based explanation methods estimate input relevance by systematically altering input regions and measuring the resulting change in the model output.
While observing input-output relationships is an intuitive approach to assess relevance, perturbation inherits a key limitation of attribution methods, namely the sensitivity to methodological settings. 
Design choices such as the size and shape of the perturbed regions, the way features are grouped, and the replacement scheme directly influence the estimated relevance, which can lead to ambiguous and even contradicting explanations.

Our work addresses this disagreement in perturbation-based xAI by providing a systematic analysis of two key parameters: first, the perturbation patch geometry, including both size and shape of the perturbed region, and second, the perturbation type, defined by the replacement scheme.
By examining how these parameters affect relevance estimation, our study demonstrates how different perturbation choices can steer the resulting explanations, and ultimately shape how model predictions are interpreted.
Our analysis combines qualitative and quantitative approaches, including visual inspection, consistency assessment, and faithfulness evaluation.
Together, these complementary perspectives provide a comprehensive understanding of how perturbation choices reflect in the resulting relevance maps and how reliably those maps capture model behavior.
Our discussion is grounded in the use case on flood detection from SAR images, where dataset specific characteristics motivate particular care in selecting perturbation schemes and in considering how such schemes may interact with dataset properties.

Overall, our findings underscore that perturbation design is not a technical detail but a determining factor in relevance estimation. 
Perturbation choices should therefore be carefully inspected, evaluated, and justified against the specific use case and dataset constraints.
Treating perturbation design as an integral part of explanation interpretation strengthens the robustness of the explanations and supports more grounded conclusions about the model predictions.

\bibliographystyle{elsarticle-harv} 
\bibliography{references.bib}

\end{document}